\documentclass[letterpaper]{article} 
\usepackage[preprint]{aaai2027}  
\usepackage[hyphens]{url}  
\usepackage{graphicx} 
\usepackage{natbib}  
\usepackage{caption} 
\usepackage{algorithm}
\usepackage{algorithmic}

\usepackage{newfloat}
\usepackage{listings}
\DeclareCaptionStyle{ruled}{labelfont=normalfont,labelsep=colon,strut=off} 
\floatstyle{ruled}
\newfloat{listing}{tb}{lst}{}
\floatname{listing}{Listing}

\usepackage{booktabs}

\usepackage{multirow}
\usepackage{tabularx}

\usepackage[table]{xcolor}
\usepackage{amsmath}
\usepackage{amssymb}

\newcolumntype{Y}{>{\centering\arraybackslash}X}
\newcolumntype{C}[1]{>{\centering\arraybackslash}p{#1}}

\title{iFAN: Inference-Aware Learning for Plain Mask Transformers}
\author{
    Fang Li\textsuperscript{\rm 1}\equalcontrib,
    Yu He\textsuperscript{\rm 1}\equalcontrib,
    Haoyang Tong\textsuperscript{\rm 1,4},
    Lichen Ma\textsuperscript{\rm 1,3},
    Jingling Fu\textsuperscript{\rm 1},
    Wenxiao Fan\textsuperscript{\rm 1,2},
    Tongxuan Liu\textsuperscript{\rm 1},
    Luohang Liu\textsuperscript{\rm 1},
    Ke Zhang\textsuperscript{\rm 1},
    Junshi Huang\textsuperscript{\rm 1}\corresponding
}
\affiliations{
    \textsuperscript{\rm 1}JD.com 
    \textsuperscript{\rm 2}Beijing Institute of Technology
    \textsuperscript{\rm 3}Xi'an Jiaotong University
    \textsuperscript{\rm 4}University of Chinese Academy of Sciences\\
    
    \tt\small \{lifang0273, heyu2579, junshi.huang\}@gmail.com
}

\newcommand{\qincnew}[1]{%
    \textsuperscript{\tiny\(\mathord{\uparrow}#1\)}%
}
\newcommand{\cmarknew}{%
    \textsubscript{c}%
}
\newcommand{\nscore}[1]{#1}
\newcommand{\qscore}[2]{%
    #1\makebox[0pt][l]{\qincnew{#2}}%
}
\newcommand{\nqscore}[2]{%
    #1\makebox[0pt][l]{\textsuperscript{\tiny #2}}%
}
\newcommand{\cscore}[1]{%
    #1\makebox[0pt][l]{\cmarknew}%
}
\newcommand{\cqscore}[2]{%
    #1%
    \makebox[0pt][l]{\qincnew{#2}}%
    \makebox[0pt][l]{\cmarknew}%
}

\begin{document}

\maketitle

\begin{abstract}

Query-based mask transformers assemble segmentation outputs through pixel-wise competition among query predictions of the final layer, yet this inference process is not explicitly optimized during training. 
We identify two key mismatches: the query with the highest probability–mask score does not necessarily produce the most accurate mask, and final-layer decoding may discard superior predictions from intermediate layers. 
To address these issues, we propose Inference-Aware Learning (\textbf{iFAN}), a general training framework for plain mask transformers. 
iFAN introduces Adjusted Probability-Mask Ranking (APMR), which aligns query competition with predicted mask quality and suppresses high-confidence but inaccurate competitors. 
We further employ Cross-Layer Self-Distillation (CLSD) to transfer stronger intermediate predictions to the final layer. 
The ranking and distillation objectives are training-only, while inference retains efficient final-layer decoding.
Experiments on COCO, ADE20K, and Cityscapes demonstrate consistent improvements across panoptic, instance, and semantic segmentation, as well as across different architectures, backbone scales, and input resolutions. 
Overall, iFAN improves performance by an average of 1.20 PQ, 1.30 AP, and 0.63 mIoU, with negligible additional parameters, FLOPs and inference latency.


\end{abstract}

\section{Introduction}

\begin{figure}[h]
  \centering
  \includegraphics[width=\linewidth]{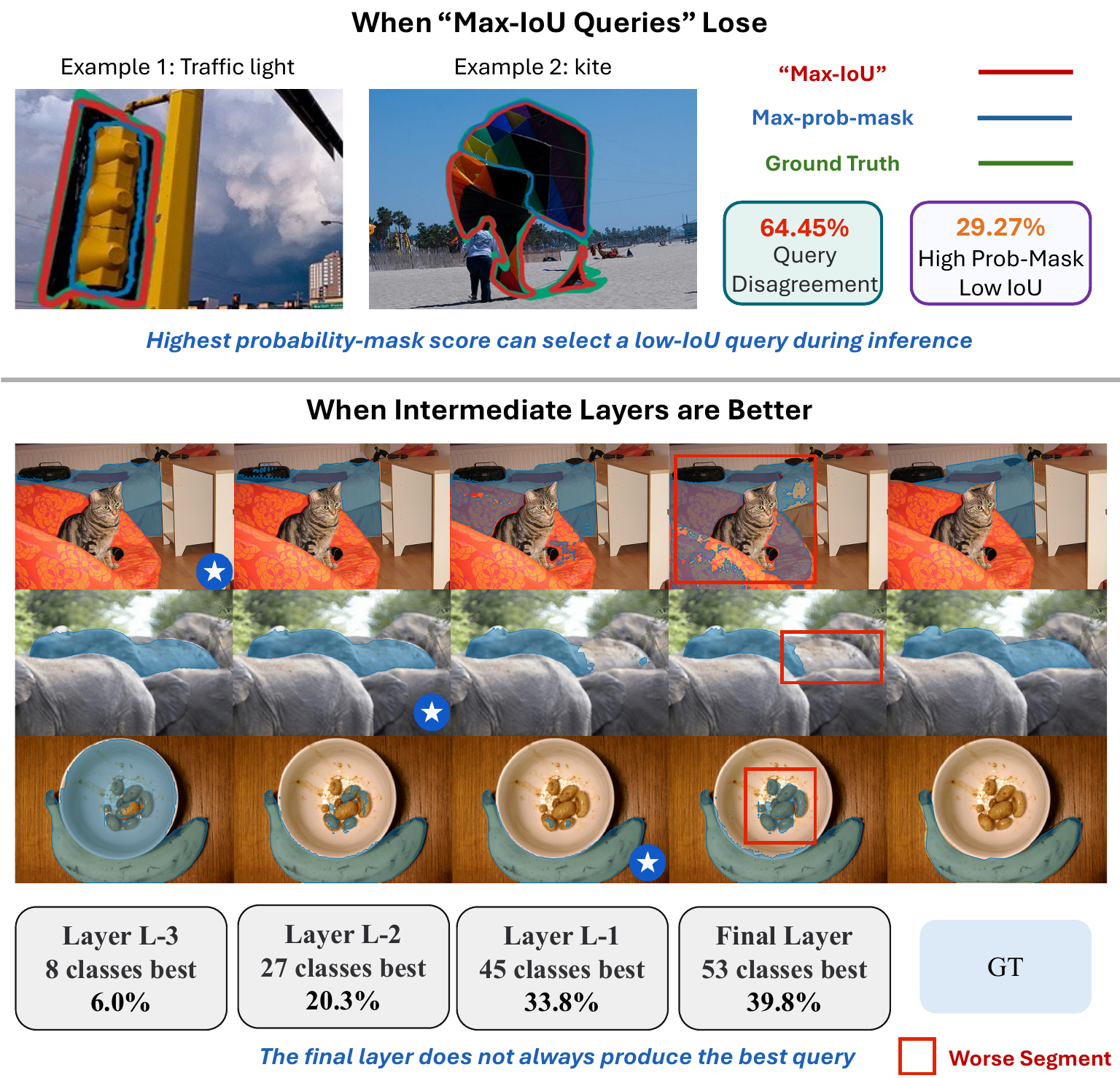}
    \caption{
    We study the performance of mask transformer on COCO val2017 among queries and layers with ground truth masks.
    Top: the \textit{max-prob-mask} query (blue) is selected to generate the mask result of target object while achieving inferior performance than the \textbf{oracle result} of ``\textit{max-IoU}'' query (red), which incurs $64.45\%$ suboptimal result.
    Bottom: the distribution of ``\textit{max-IoU}'' query across layers. A better query can be found in intermediate layer (star indicates max IoU), which can provide self-distillation information.
    }
  \label{fig:intro}
\end{figure}

Query-based segmentation has become a promising framework for semantic, instance, and panoptic segmentation~\citep{wang2021max,yu2022k,jain2023oneformer,Shahabodini_2026_CVPR}. 
Unlike conventional approaches that classify each pixel using a fixed set of prediction headers, modern mask transformers segment an image with a set of learnable queries. 
Each query independently predicts a class distribution together with a corresponding mask, and the final segmentation result is obtained by a competition among these query-wise prediction maps during output assembly~\citep{li2023mask}. 
This query-based formulation is naturally compatible with Vision Transformers (ViTs)~\citep{dosovitskiy2020vit}, since image tokens and query tokens can be processed within the same attention framework. 
Recent encoder-only methods further show that, based on strong pre-training backbones such as DINOv2~\citep{oquab2024dinov2}, a plain ViT can support segmentation by simply injecting queries into its later layers without bells and whistles~\citep{kerssies2025your}.

In this paper, we rethink the inference process of mask transformers and identify the limitations of widely used max-prob-mask strategy~\cite{cheng2021per}.
In this strategy, each query predicts a class distribution and a class-agnostic foreground probability mask, which are then combined to determine the pixel-wise segments.
However, the mask predicted by each query is isolated with class prediction and thus lacks global information of object, \textit{e.g.}, the object shape and contour~\citep{ke2022masktransfiner}. 
Consequently, the predicted class may be inconsistent with the predicted mask~\citep{hu2023pais}. 
For example, in the \textit{traffic-light} case of Fig. \ref{fig:intro}, although the max-prob-mask strategy predicts the correct category, the resulting segmentation covers only the most salient regions of the traffic light. 
For the object without a consistent shape, such as \textit{kite} in Fig. \ref{fig:intro}, query-based models hardly predict the whole region.
Based on the inference results of a mask transformer~\citep{kerssies2025your} on COCO val2017~\citep{lin2014microsoft}, we find that 64.45\% of the segmentation results selected by max-prob-mask strategy are different from their theoretical upper-bound counterparts by ``max-IoU'' strategy. 
With a strict threshold, 29.27\% of the results exhibit high prob-mask scores while low IoUs.
 
We claim that these suboptimal results are induced by the incomplete competition strategy of probability-mask pairs \citep{cheng2021per}.
To address this issue, we introduce a global mask quality score for each query to adjust the probability-mask score map for segmentation prediction. 
However, mask-quality calibration alone cannot fully suppress high-scoring yet inaccurate competitors.
Hungarian matching determines the optimal assignment at each iteration but does not preserve query identities across training: a query may switch repeatedly between the states of matched foreground and unmatched background~\citep{li2022dn,liu2023detection,zhang2022dino}. 
Importantly, unmatched queries are supervised only by no-object classification, without explicit suppression of their mask responses~\citep{cheng2022masked}. 
They may retain plausible knowledge of foreground mask from previous iterations, causing multiple plausible masks for the same object. 
These plausible masks pose strong competition during inference, motivating us to introduce an explicit cross-query ranking objective~\citep{pu2023rankdetr}.
Specifically, for each foreground target, we identify the unmatched queries with high probability of target class yet low mask overlap as hard negatives.
We then introduce a pairwise ranking objective that encourages the matched query to produce a higher probability-mask score than that of negatives. 
To stabilize the learning of classification and mask prediction, we only use the gradient of ranking loss \textit{w.r.t} mask quality to update the model parameters. 

Most mask transformers use only the output of final layer to produce segmentation predictions, even if the auxiliary supervision is provided at intermediate layers.
We analyze the segmentation results generated by the last several layers of a mask transformer~\citep{kerssies2025your}. 
As shown in Figure~\ref{fig:intro}, only 39.8\% of ground-truth targets obtain their best predictions at the final layer, while the penultimate layer achieves comparable overall performance. 
Simply aggregating the predictions of multiple layers can improve performance, but it also introduces additional post-processing complexity and computational costs. 
Instead, we propose to extract high-quality information from preceding layers to supervise the final layer while keeping the inference strategy unchanged. 
Therefore, the Cross-Layer Self-Distillation (CLSD) is integrated in the training stage to distill high-quality intermediate-layer information by the final layer while keeping the inference stage unchanged~\citep{zhang2019byot,wang2024kddetr}.

In summary, we upgrade the training framework of mask transformers motivated by the deficiencies of segmentation inference, which thus refers to as inference-aware learning task.
Our main contributions are summarized as follows:

\begin{itemize}
  \item We adjust the probability-mask scoring criterion with a global quality score to address the deficiency of inference. 
  In addition, we employ object-oriented ranking task to produce more discriminative prediction scores.
  \item We fully exploit the useful information contained in different layers of mask transformers and propose a self-distillation strategy that improves the predictions of the final layer without additional inference cost.
  \item Across multiple mask-transformer architectures and datasets, iFAN consistently improves panoptic, instance, and semantic segmentation, achieving average gains of 1.20 PQ, 1.30 AP, and 0.63 mIoU, respectively, with virtually no additional inference overhead.
\end{itemize}

\section{Related Work}

\subsection{Query-Based and Quality-Aware Segmentation}

DETR introduced object queries and bipartite set prediction for detection~\citep{carion2020end}. MaskFormer and Mask2Former extend this formulation to segmentation through class--mask prediction~\citep{cheng2021per,cheng2022masked}, while OneFormer and Mask DINO develop it for universal segmentation and joint detection--segmentation modeling~\citep{jain2023oneformer,li2023mask}. 
EoMT and PMT further inserts segmentation queries directly into the late blocks of a pre-trained ViT~\citep{kerssies2025your,cavagnero2026pmt}. 
Despite their architectural differences, these methods must assemble unordered class--mask predictions into a final output, making the reliability of the probability-mask score central to query competition.

Quality estimation and ranking improve prediction reliability. 
Mask Scoring R-CNN calibrates instance scores with predicted mask IoU~\citep{huang2019mask}; Panoptic SegFormer combines classification and segmentation quality to resolve mask overlaps~\citep{li2022panoptic}; and Rank-DETR learns quality-aware ranking for detection queries~\citep{pu2023rankdetr}. 
These methods focus on instance-score calibration, inference-time overlap handling, or box-level ranking. 
Adjusted Probability-Mask Ranking (APMR) instead optimizes the pixel-wise competition of mask-transformer inference. 
It learns a soft-IoU-supervised mask-quality score, constructs an adjusted probability-mask score map, and ranks each matched query above hard unmatched competitors within its target region. 
Its ranking gradients update only the quality branch, preserving the original classification and mask objectives.

\subsection{Intermediate Supervision and Self-Distillation}

Intermediate supervision applies annotation-based objectives before the final output. 
Deeply-Supervised Nets introduced auxiliary hidden-layer losses~\citep{lee2015deeply}, while query-based segmenters supervise intermediate class and mask predictions~\citep{li2022panoptic}. 
MP-Former further promotes mask consistency across decoder layers~\citep{zhang2023mpformer}. These approaches improve intermediate predictions or their evolution, but do not explicitly transfer a better intermediate probability-mask prediction to the final layer.

Self-distillation transfers knowledge between predictions within a training process, as exemplified by Be Your Own Teacher~\citep{zhang2019byot}. For set prediction, DETRDistill uses a pre-trained teacher to transfer matched logits, features, and query priors, whereas OD-DETR uses an exponential-moving-average teacher~\citep{chang2023detrdistill,wu2024oddetr}. Cross-Layer Self-Distillation (CLSD) requires neither. 
For each target, it selects the intermediate matched query with the highest soft-IoU and retains it only when it outperforms the final prediction. 
CLSD then distills its adjusted probability-mask score map $\hat{S}$ through a one-sided objective, transferring useful intermediate knowledge while retaining final-layer-only inference.

\section{Method}

\begin{figure*}[t]
  \centering
  \includegraphics[width=\linewidth]{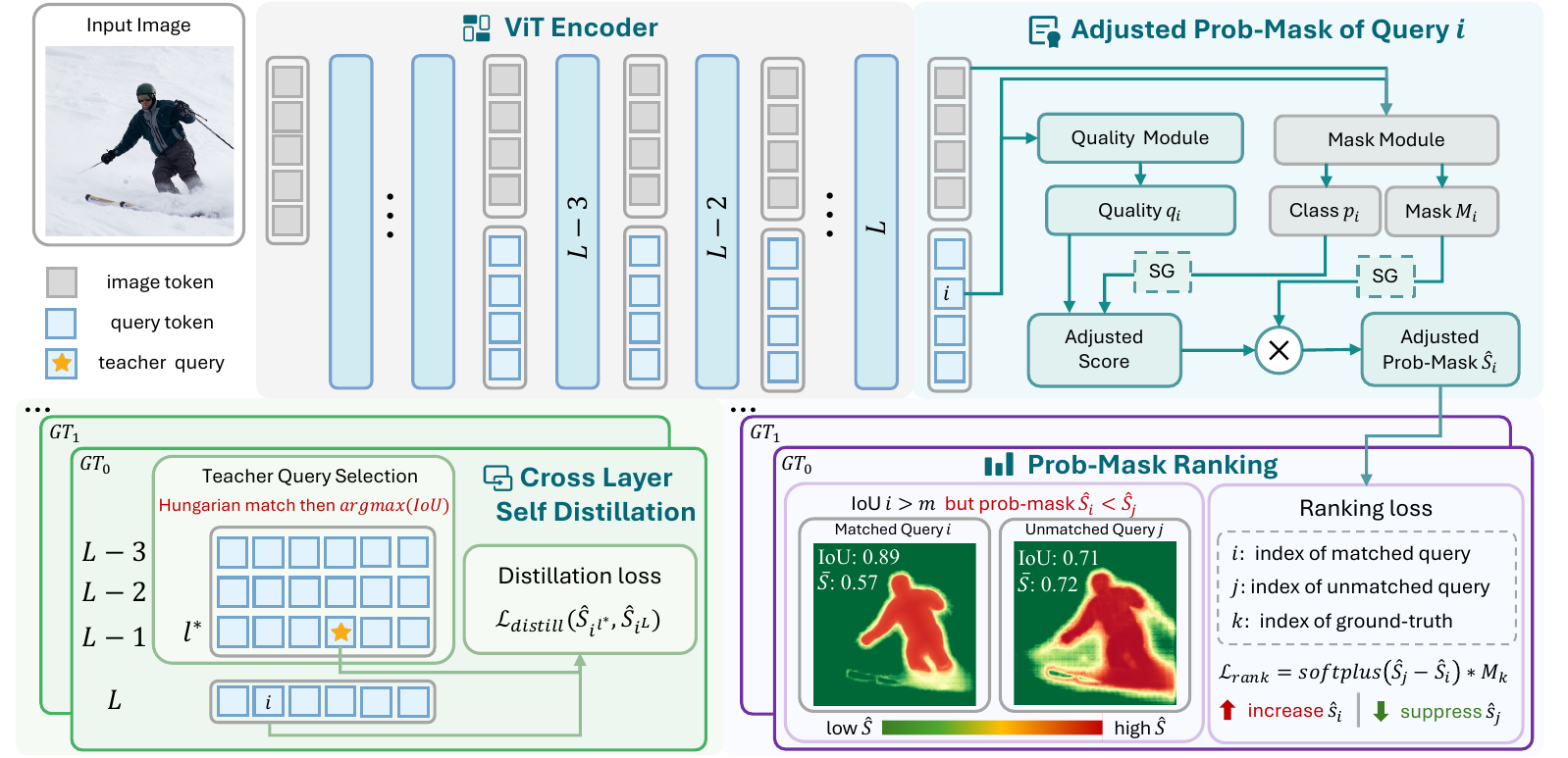}
  \caption{\textbf{Overview of iFAN.}
  APMR addresses the query-wise mismatch by learning mask quality, calibrating the scores used for assembly, and ranking a matched query above high-evidence unmatched query.
  CLSD addresses the layer-wise mismatch by selecting a stronger teacher query from an intermediate layer and transferring its assembled evidence to the final layer.
  At inference, the adjusted-prob-mask is only computed for final layer. 
  The APMR ranking objective and CLSD are training-only.}
  \label{fig:framework}
\end{figure*}

In this section, we introduce the details of \textbf{Inference-Aware Learning (iFAN)} as illustrated in Fig.~\ref{fig:framework}.
Based on the plain mask transformer, we adjust the predicted probability-mask map by mask quality and Probability-Mask Ranking strategy, promoting the distinction of final predictions from matched queries and unmatched queries.
To further acquire the knowledge of intermediate layers, the Cross-Layer Self-Distillation is injected into our framework for the promotion of matched queries.
Those strategies motivated by the deficiencies in inference stage improve the performance of mask transformer, composing the framework of Inference-Aware Learning.

\subsection{Preliminaries}

Given an input image $I \in R^{H \times W}$, the mask transformer predicts $N$ probability-mask pairs
\begin{equation}
\hat{y} = \{ (\hat{\mathbf{p}}_i, \hat{M}_i) \}^{N}_{i=1}
\end{equation}
where the pair $(\hat{\mathbf{p}}_i, \hat{M}_i) $ is the prediction result from $i$-th query.
$\mathbf{\hat{p}}_i\in[0,1]^{K+1}$ is the class-probability distribution over $K$ foreground classes and the no-object class $\varnothing$.
$\hat{M}_i$ is the mask probability map of the foreground, which is usually class-agnostic.
Similarly, the ground-truth segments are represented as
\begin{equation}
y = \{(c_i , M_i )| c_i \in \{1, \dots, K\}, M_i \in \{0, 1\}^{H \times W} \}^{N^{gt}}_{i=1}
\end{equation}

During training, the Hungarian matching algorithm assigns each ground-truth to a unique query by minimizing the matching cost between pairs of query predictions and ground-truths.
The learning tasks on the matched pairs include object classification and foreground mask prediction.
Unmatched queries are usually considered as background with only the supervision of no-object classification.

The inference stage follows a slightly different decision rule. 
Since queries may predict overlapping masks of objects, the segmentation assembly strategy must assign each pixel to one of the $N$ predicted probability-mask pairs according to the maximal probability-mask score of object class and object-agnostic mask.

\subsection{Adjusted Probability-Mask Ranking}
\label{sec:CQR}

The standard segmentation assembly constructs the segmentation result based on the pixel-wise competition among probability-mask responses, without considering the holistic reliability of the predicted mask. 
Therefore, we introduce a lightweight shared quality head to estimate the global mask quality $\hat{q}_i$ of the $i$-th query, resulting in the prediction tuple $(\hat{q}_i, \hat{\mathbf{p}}_i, \hat{M}_i)$. The quality head consists of a single linear layer followed by a sigmoid function, and its parameters are shared across all queries and prediction layers, introducing only negligible parameter overhead.
The ground-truth of $\hat{q}_i$ for matched queries is the \textit{soft\_IoU} of predicted mask $\hat{M}_i$ and its corresponding ground-truth segmentation mask, and is set as $0$ for unmatched queries.
In the training stage, the BCE loss is integrated into the segmentation task for the learning of mask quality, comprising the augmented segmentation loss, and the unmatched queries with high class probability are selected as hard negatives. 

To complement global information in segmentation competition, we integrate predicted quality into probability-mask and obtain the adjusted probability-mask pair $((\hat{q}_i * \hat{\mathbf{p}}_i)^\gamma, \hat{M}_i)$, where $\gamma$ is set to 2 to enhance discrimination in the high-score range.
Together with the hard mining strategy, the adjusted score $(\hat{q}_i * \hat{\mathbf{p}}_i)^\gamma$ can suppress the plausible result of unmatched queries induced by modest predicted masks.

With the adjusted probability-mask pair, we introduce an object-oriented ranking objective to directly model the segmentation competition between matched and unmatched queries.
This objective requires that the adjusted probability-mask score of matched query should be notably higher than that of unmatched queries over the foreground region of target. 
Given target object $(c_k , M_k)$ and the adjusted probability-mask pair of matched query $((\hat{q}_i * \hat{\mathbf{p}}_i)^\gamma, \hat{M}_i)$, the probability-mask score map of $i$-th query for $k$-th target object is formulated as
\begin{equation}
\hat{S}_i =
(\hat{q}_i * \hat{\mathbf{p}}_i(c_k))^\gamma * \hat{M}_i
\end{equation}
where $\hat{\mathbf{p}}_i(c_k)$ is the $c_k$-th value of $\hat{\mathbf{p}}_i$.
Similarly, denoting by the probability-mask score map of unmatched query for $k$-th target object $\hat{S}_j = ((\hat{q}_j * \hat{\mathbf{p}}_j(c_k))^\gamma, \hat{M}_j)$, the ranking loss between the score maps of matched query $\hat{S}_i$ and unmatched query $\hat{S}_j$ with respect to $k$-th target object is formulated as:

\begin{equation}
\mathcal{L'}_{\mathrm{rank}}=
\frac{1}{|M_k|}
\sum_{}
(
\mathrm{\textit{softplus}}
\left(
\hat{S}_j - \hat{S}_i
\right)
* M_k
),
\end{equation}
Then, the overall ranking loss $\mathcal{L}_{\mathrm{rank}}$ is the average of $\mathcal{L'}_{\mathrm{rank}}$ over all target objects and their corresponding hard negatives.
In our implementation, we select the top $N_H$ unmatched queries according to the order of $\hat{S}_j$.
Softplus provides smooth gradients to continuously enlarge the score gap between matched queries and hard negatives. 
The gradients with respect to $\hat{\mathbf{p}}$ and $\hat{M}$ are detached to prevent over-suppression of the class probabilities and mask predictions.

\subsection{Cross-Layer Self-Distillation}
\label{sec:CLD}

During training, predictions from multiple layers receive auxiliary supervision, whereas inference only retains the prediction of final layer.
Although cascaded supervision allows different layers to capture complementary information, valuable intermediate representations may not be fully preserved in later layers\citep{liu2023detection,zhang2023mpformer}, as illustrated in Fig.~\ref{fig:intro}.
Therefore, we propose the strategy of Cross-Layer Self-Distillation (CLSD) to extract high-quality intermediate knowledge as teacher supervision for the final layer.

For a target object $(c_k , M_k)$, the Hungarian matching algorithm is applied at each $\ell$ layer, where $\ell \in [L-4, L]$ and $L$ is the index of final layer.
Let $i^\ell$ denotes the index of matched query at $\ell$-th layer, which may be different in various layers.
Among all matched queries in intermediate layers, we select the layer with the highest soft IoU as the teacher candidate:
\begin{equation}
\ell^\star=
\mathop{\mathrm{arg\,max}}_{L-3 \le \ell < L} \operatorname{\textit{soft\_IoU}(}M_k, \hat{M}_{i^{\ell}}).
\label{eq:teacher_selection}
\end{equation}
Considering that $i^{\ell^*}$-th query (at $\ell^*$-layer) and $i^L$-th query (at $L$-layer) are all matched queries of target object $(c_k , M_k)$ at different layers,  we select the real teacher supervision if the $\operatorname{\textit{soft\_IoU}(}M_k, \hat{M}_{i^{\ell^*}})$ of $i^{\ell^*}$-th query (at $\ell^*$-layer) is better than that of $i^L$-th query (at $L$-layer).
In this way, we design the loss function of Cross-Layer Self-Distillation for target object $(c_k , M_k)$:
\begin{equation}
\mathcal{L'}_{\mathrm{distill}}
=
\frac{1}{|M_k|}
\displaystyle
\sum_{}
\left[
(
\operatorname{\textit{sg}}(\hat{S}_{i^{\ell^*}}) - \hat{S}_{i^L}
) * M_k
\right]_+ 
\end{equation}
where $[\cdot]_+=\max(0,\cdot)$ and $\operatorname{\textit{sg}}(\cdot)$ denotes the stop-gradient operator.
The hinge function is activated when the ``soft\_IoU-qualified'' teacher provides stronger adjusted probability-mask score map than that of final layer. 
Analogously, the overall distillation loss $\mathcal{L}_{\mathrm{distill}}$ is average of $\mathcal{L'}_{\mathrm{distil
l}}$ over all target objects.
Note that CLSD applied on training stage brings no additional cost in inference.

\begin{table*}[t]
    \centering
    {
    
    \fontsize{9pt}{10.8pt}\selectfont
    \rmfamily
    \setlength{\tabcolsep}{1mm}
    \renewcommand{\arraystretch}{1.0}
    \begin{tabularx}{\linewidth}{lllc l YYYC{14mm} l YYYY}
        \toprule
        \multirow{2}{*}{Method} &
        \multirow{2}{*}{Backbone} &
        \multirow{2}{*}{Pre-training} &
        \multirow{2}{*}{Params} &&
        \multicolumn{4}{c}{COCO \emph{val2017}~\cite{lin2014microsoft}} &&
        \multicolumn{4}{c}{ADE20K \emph{val}~\cite{zhou2017ade20k}} \\
        \cmidrule{6-9}\cmidrule{11-14}
        &&&&& Input & GFLOPs & FPS & PQ &&
        Input & GFLOPs & FPS & PQ \\
        \midrule

        Mask2Former$^\dagger$ & Swin-L & IN21K & 216M &&
        $800^2$ & 868 & 24 & \nscore{57.8} &&
        $640^2$ & -- & 33 & \nscore{48.1} \\

        kMaX-DeepLab & ConvNeXt-L & IN21K & 232M &&
        $1281^2$ & -- & -- & \nscore{58.0} &&
        $1281^2$ & 1302 & -- & \nscore{50.9} \\

        OneFormer$^\dagger$ & DiNAT-L & IN21K & 223M &&
        $800^2$ & 736 & 20 & \nscore{58.0} &&
        $1280^2$ & 1369 & 10 & \cscore{53.5} \\

        MaskDINO$^\dagger$ & Swin-L & IN21K & 223M &&
        $800^2$ & 1326 & 14 & \nscore{58.3} &&
        -- & -- & -- & -- \\

        OneFormer+ViT-P$^\dagger$ & DiNAT-L & IN21K & 309M &&
        $1024^2$ & 1013 & 8 & \nscore{58.0} &&
        $1280^2$ & 1955 & 3 & \cscore{54.0} \\

        \midrule

        Mask2Former$^\ddagger$ &
        ViT-AL$^\ddagger$ & DINOv2 & 349M &&
        $640^2$ & 830 & 29 & \nscore{57.1} &&
        $640^2$ & 830 & 29 & \cscore{51.8} \\

        Mask2Former$^\ddagger$ &
        ViT-AL$^\ddagger$ & DINOv2 & 354M &&
        $1280^2$ & 4817 & 10 & \nscore{59.7} &&
        $1280^2$ & 4817 & 10 & \cscore{53.0} \\

        Mask2Former$^\ddagger$ &
        ViT-Adapter-g$^\ddagger$ & DINOv2 & 1209M &&
        $640^2$ & 2510 & 20 & \nscore{57.7} &&
        $640^2$ & 2510 & 20 & \cscore{52.6} \\

        Mask2Former$^\ddagger$ &
        ViT-Adapter-g$^\ddagger$ & DINOv2 & 1216M &&
        $1280^2$ & 13790 & 6 & \nscore{59.9} &&
        $1280^2$ & 13790 & 6 & \cscore{54.2} \\

        \midrule

        PMT & ViT-L & DINOv3 & 357M &&
        $640^2$ & 767 & 141 & \nscore{56.1} &&
        $640^2$ & 767 & 141 & \nqscore{49.4}{$^\ddagger$} \\
         
        \rowcolor[RGB]{230,245,255}
        PMT-iFAN & ViT-L & DINOv3 & 357M &&
        $640^2$ & 767 & 141 & \qscore{56.7}{0.6} &&
        $640^2$ & 767 & 141 & \qscore{50.7}{1.3} \\

        PMT & ViT-L & DINOv3 & 357M &&
        $1280^2$ & 4925 & 29 & \nscore{58.1} &&
        $1280^2$ & 4925 & 29 & \nqscore{50.5}{$^\ddagger$} \\

        \rowcolor[RGB]{230,245,255}
        PMT-iFAN & ViT-L & DINOv3 & 357M &&
        $1280^2$ & 4925 & 29 & \qscore{58.6}{0.5} &&
        $1280^2$ & 4925 & 29 & \qscore{53.0}{3.0} \\

        \midrule

        EoMT & ViT-L & DINOv2 & 316M &&
        $640^2$ & 669 & 128 & \nscore{56.0} &&
        $640^2$ & 669 & 128 & \cscore{50.6} \\
        
        \rowcolor[RGB]{230,245,255}
        EoMT-iFAN & ViT-L & DINOv2 & 316M &&
        $640^2$ & 669 & 128 & \qscore{57.0}{1.0} &&
        $640^2$ & 669 & 128 & \cqscore{52.2}{1.6} \\

        EoMT & ViT-L & DINOv2 & 322M &&
        $1280^2$ & 4146 & 30 & \nscore{58.3} &&
        $1280^2$ & 4146 & 30 & \cscore{51.7} \\
        
        \rowcolor[RGB]{230,245,255}
        EoMT-iFAN & ViT-L & DINOv2 & 322M &&
        $1280^2$ & 4146 & 30 & \qscore{58.8}{0.5} &&
        $1280^2$ & 4146 & 30 & \cqscore{53.9}{2.2} \\

        EoMT & ViT-G & DINOv2 & 1164M &&
        $640^2$ & 2261 & 55 & \nscore{57.0} &&
        $640^2$ & 2261 & 55 & \cscore{51.3} \\
        
        \rowcolor[RGB]{230,245,255}
        EoMT-iFAN & ViT-G & DINOv2 & 1164M &&
        $640^2$ & 2261 & 55 & \qscore{57.8}{0.8} &&
        $640^2$ & 2261 & 55 & \cqscore{52.7}{1.4} \\

        EoMT & ViT-G & DINOv2 & 1171M &&
        $1280^2$ & 12712 & 12 & \nscore{59.2} &&
        $1280^2$ & 12712 & 12 & \cscore{52.8} \\
        
        \rowcolor[RGB]{230,245,255}
        EoMT-iFAN & ViT-G & DINOv2 & 1171M &&
        $1280^2$ & 12712 & 12 & \qscore{59.6}{0.4} &&
        $1280^2$ & 12712 & 12 & \cqscore{53.9}{1.1} \\

        \bottomrule
    \end{tabularx}
    }

    \caption{\textbf{Panoptic segmentation on COCO \emph{val2017} and
    ADE20K \emph{val}.} Blue rows apply iFAN to the corresponding PMT
    or EoMT baseline; $\uparrow$ gives the absolute PQ improvement.
    $^\dagger$ denotes shortest-side resizing, $^\ddagger$
    re-implementation, and \textsubscript{c} COCO panoptic
    pre-training. Parameters, GFLOPs, and FPS in iFAN rows follow the
    matched baseline under shared-configuration accounting rather than
    an independent efficiency measurement.}
    \label{tab:main_panoptic}
\end{table*}

\subsection{Training and Inference}
The overall objective combines the augmented segmentation loss with adjusted probability-mask ranking loss and self-distillation loss
\begin{equation}
\mathcal{L}=
\mathcal{L}_{\mathrm{aug\_seg}}
+\lambda_{\mathrm{rank}}\mathcal{L}_{\mathrm{rank}}
+\lambda_{\mathrm{distill}}\mathcal{L}_{\mathrm{distill}}.
\end{equation}
where $\mathcal{L}_{\mathrm{aug\_seg}}$ is the augmented segmentation loss, including object classification, pixel-wise mask prediction, Dice and proposed mask-quality regression losses. 
The coefficients $\lambda_{\mathrm{rank}}$ and $\lambda_{\mathrm{distill}}$ balance the contributions of probability-mask score ranking and cross-layer self-distillation objectives, which are set to 0.1 and 0.4, respectively. 
The epoch-dependent coefficient $\lambda_{\mathrm{distill}}$ gradually activates selective depth distillation after an initial warm-up stage and anneals near convergence stage, mitigating the influence of unreliable early teachers and avoiding excessive constraints on the final representation. 
During inference, we extract the adjusted probability-mask score map $\hat{S}_i = (\hat{q_i} * \mathrm{\textit{max}}(\hat{\mathbf{p_i}}))^\gamma * \hat{M_i}$ for $i$-th query at the final layer  for segmentation competition.

\section{Experiments}
\label{sec:exp}

\subsection{Experimental Setup}

\paragraph{Datasets and metrics.}
We evaluate iFAN on panoptic, instance, and semantic segmentation.
COCO val2017~\citep{lin2014microsoft} is used for panoptic and instance segmentation, ADE20K val~\citep{zhou2017ade20k} for panoptic and semantic segmentation, and Cityscapes val~\citep{cordts2016cityscapes} for semantic segmentation.
Following the evaluation protocols, we report Panoptic Quality (PQ)~\citep{kirillov2019panoptic}, mask Average Precision (AP)~\citep{lin2014microsoft}, and mean Intersection over Union (mIoU)~\citep{everingham2010pascal}, respectively.

\paragraph{Implementation}
We implement iFAN on two plain mask-transformer baselines: EoMT~\citep{kerssies2025your} uses DINOv2-pretrained ViT-S/B/L/G backbones, while PMT~\citep{cavagnero2026pmt} uses DINOv3-pretrained ViT-L.
Models are trained on two NVIDIA B200 GPUs with a batch size of 4, using mixed-precision training and AdamW~\citep{loshchilov2017decoupled}. 
We use an initial learning rate of $10^{-4}$, layer-wise learning-rate decay of 0.8, and polynomial learning-rate decay with a power of 0.9. 
Training lasts 12 epochs on COCO, 31 epochs on ADE20K, and 56 epochs on Cityscapes. 
We set the number of hard negatives $N_H$ to 10.
CLSD starts at epoch 4 and is linearly annealed thereafter.
Unless stated otherwise, all ablations use EoMT-iFAN with ViT-L and $640^2$ inputs on COCO.

\subsection{Main Results}

\begin{table*}[t]
    \centering
    {
    \fontsize{9pt}{10.8pt}\selectfont
    \rmfamily
    \setlength{\tabcolsep}{1mm}
    \renewcommand{\arraystretch}{1.0}
    \begin{tabularx}{\linewidth}{lllc c cYYC{13mm} l YYYY}
        \toprule
        \multirow{2}{*}{Method} &
        \multirow{2}{*}{Backbone} &
        \multirow{2}{*}{Pre-training} &
        \multirow{2}{*}{Params} &&
        \multicolumn{4}{c}{
            Cityscapes \emph{val}~\cite{cordts2016cityscapes}
        } &&
        \multicolumn{4}{c}{
            ADE20K \emph{val}~\cite{zhou2017ade20k}
        } \\
        \cmidrule{6-9}\cmidrule{11-14}
        &&&&&
        Input & GFLOPs & FPS & mIoU &&
        Input & GFLOPs & FPS & mIoU \\
        \midrule

        Mask2Former$^\dagger$ & Swin-L & IN21K & 216M &&
        $1024\times2048$ & -- & 14 & \nscore{83.3} &&
        $640^2$ & -- & 33 & \nscore{56.1} \\

        MaskDINO$^\dagger$ & Swin-L & IN21K & 223M &&
        -- & -- & -- & -- &&
        $640^2$ & -- & -- & \nscore{56.6} \\

        OneFormer$^\dagger$ & ConvNeXt-XL & IN21K & 373M &&
        $1024\times2048$ & 775 & 7 & \nscore{83.6} &&
        $640^2$ & 607 & 21 & \nscore{57.4} \\

        OneFormer$^\dagger$ & DiNAT-L & IN21K & 223M &&
        $1024\times2048$ & 450 & 14 & \nscore{83.1} &&
        $896^2$ & 678 & 19 & \nscore{58.1} \\

        kMaX-DeepLab & ConvNeXt-L & IN21K & 232M &&
        $1025\times2049$ & 1673 & -- & \nscore{83.5} &&
        -- & -- & -- & -- \\

        Mask2Former & ViT-L & DINOv2 & -- &&
        $896\times1792$ & -- & -- & \nscore{84.8} &&
        $896^2$ & -- & -- & \nscore{59.4} \\

        OneFormer+ViT-P$^\ddagger$ &
        DiNAT-L$^\ddagger$ & IN21K & 309M &&
        -- & -- & -- & -- &&
        $1280^2$ & 1955 & 5 & \cscore{59.9} \\

        \midrule

        Mask2Former$^\ddagger$ &
        ViT-AL$^\ddagger$ & DINOv2 & 351M &&
        $1024^2$ & 5200 & 7 & \nscore{84.5} &&
        $512^2$ & 910 & 21 & \nscore{58.9} \\

        \midrule
        
        PMT & ViT-L & DINOv3 & 357M &&
        $1024^2$ & 5113 & 27 & \nqscore{83.1}{$^\ddagger$} &&
        $512^2$ & 823 & 128 & \nscore{58.5} \\
        
        \rowcolor[RGB]{230,245,255}
        PMT-iFAN & ViT-L & DINOv3 & 357M &&
        $1024^2$ & 5113 & 27 & \qscore{83.8}{0.7} &&
        $512^2$ & 823 & 128 & \qscore{59.4}{0.9} \\
        
        \midrule

        EoMT & ViT-L & DINOv2 & 319M &&
        $1024^2$ & 4350 & 25 & \nscore{84.2} &&
        $512^2$ & 721 & 92 & \nscore{58.4} \\
        
        \rowcolor[RGB]{230,245,255}
        EoMT-iFAN & ViT-L & DINOv2 & 319M &&
        $1024^2$ & 4350 & 25 & \qscore{84.5}{0.3} &&
        $512^2$ & 721 & 92 & \qscore{59.0}{0.6} \\

        \bottomrule
    \end{tabularx}
    
    }
    \caption{\textbf{Semantic segmentation on Cityscapes and ADE20K
    \emph{val}.} Blue rows apply iFAN to the corresponding PMT or EoMT
    baseline, and $\uparrow$ gives the absolute mIoU improvement.
    $^\dagger$ denotes ADE20K shortest-side resizing, $^\ddagger$
    re-implementation, and \textsubscript{c} COCO panoptic
    pre-training. PMT and EoMT use DINOv3- and DINOv2-pretrained
    backbones, respectively. Efficiency for windowed methods includes
    all crops; iFAN rows follow the matched baseline under
    shared-configuration accounting.}
    \label{tab:main_semantic}
\end{table*}

\newcommand{\apscore}[1]{%
    #1\phantom{\qincnew{0.0}}%
}

\newcommand{\apgain}[2]{%
    #1\qincnew{#2}%
}
\begin{table}[!t]
    \centering
    {
    \fontsize{9pt}{10.8pt}\selectfont
    \rmfamily
    \setlength{\tabcolsep}{1mm}
    \renewcommand{\arraystretch}{1.0}

    \begin{tabularx}{\linewidth}{
        l l
        c c c c
    }
        \toprule
        Method & Backbone & Params & Input & FPS & \apscore{AP} \\
        \midrule
        
        OneFormer$^\dagger$ & DiNAT-L & 223M &
        $800^2$ & 20 & \apscore{49.2} \\
        
        Mask2Former$^\dagger$ & Swin-L & 216M &
        $800^2$ & 24 & \apscore{50.1} \\
        
        MaskDINO$^\dagger$ & DiNAT-L & 223M &
        $800^2$ & 14 & \apscore{52.3} \\
        
        OneFormer+ViT-P$^\dagger$ & DiNAT-L & 309M &
        $1024^2$ & 4 & \apscore{49.5} \\
        
        \midrule
        
        Mask2Former$^\ddagger$ & ViT-AL$^\ddagger$ & 349M &
        $640^2$ & 29 & \apscore{47.6} \\
        
        Mask2Former$^\ddagger$ & ViT-AL$^\ddagger$ & 354M &
        $1280^2$ & 10 & \apscore{51.4} \\
        
        \midrule
        PMT & ViT-L & 357M &
        $640^2$ & 141 & \apscore{45.4} \\
        
        \rowcolor[RGB]{230,245,255}          
        PMT-iFAN & ViT-L & 357M &
        $640^2$ & 141 & \apgain{46.6}{1.2} \\
         
        PMT & ViT-L & 357M &
        $1280^2$ & 29 & \apscore{48.8} \\
        
        \rowcolor[RGB]{230,245,255}        
        PMT-iFAN & ViT-L & 357M &
        $1280^2$ & 29 & \apgain{50.6}{1.8} \\
        
        \midrule
        
        EoMT & ViT-L & 316M &
        $640^2$ & 128 & \apscore{44.8} \\
        
        \rowcolor[RGB]{230,245,255}        
        EoMT-iFAN & ViT-L & 316M &
        $640^2$ & 128 & \apgain{46.0}{1.2} \\
        
        EoMT & ViT-L & 322M &
        $1280^2$ & 30 & \apscore{48.8} \\
        
        \rowcolor[RGB]{230,245,255}        
        EoMT-iFAN & ViT-L & 322M &
        $1280^2$ & 30 & \apgain{50.6}{1.8} \\

        \bottomrule
    \end{tabularx}
    }

    \caption{\textbf{Instance segmentation on COCO
    \emph{val2017}.} Blue rows apply iFAN to the corresponding PMT or
    EoMT baseline, and $\uparrow$ gives the absolute AP improvement.
    $^\dagger$ denotes shortest-side resizing and $^\ddagger$
    re-implementation. PMT and EoMT use DINOv3- and
    DINOv2-pretrained backbones, respectively. Parameters and FPS in
    iFAN rows follow the matched baseline under shared-configuration accounting.}
    \label{tab:main_instance}
\end{table}

\paragraph{iFAN on different benchmarks.}

As shown in Table~\ref{tab:main_panoptic}, iFAN consistently improves plain mask transformers with virtually no additional parameters, FLOPs, or inference overhead. On average, it improves PMT by 0.55 PQ on COCO and 2.15 PQ on ADE20K, and improves EoMT by 0.68 PQ and 1.58 PQ, respectively. 
Compared with more complex methods, iFAN also provides a strong accuracy--efficiency trade-off. PMT-iFAN achieves 141 FPS at $(640^2)$ resolution, while the strongest EoMT-iFAN reaches 59.6 PQ on COCO and 53.9 PQ on ADE20K, only 0.3 PQ below the best reported results. 
These results show that iFAN approaches state-of-the-art performance while preserving the efficiency of plain mask transformers.

As shown in Table~\ref{tab:main_instance}, iFAN consistently improves instance segmentation with virtually no additional inference overhead. On COCO, it improves PMT by an average of 1.1 AP and EoMT by 1.5 AP, with the strongest EoMT result increasing from 48.8 to 50.6 AP. These results demonstrate that iFAN effectively enhances instance-level recognition across different plain mask-transformer architectures.

As shown in Table~\ref{tab:main_semantic}, iFAN also yields consistent gains in semantic segmentation. It improves EoMT by 0.3 mIoU on Cityscapes and 0.6 mIoU on ADE20K, while improving PMT by 0.9 mIoU on ADE20K and 0.7 mIoU on Cityscapes. These improvements show that iFAN generalizes well across segmentation tasks, model architectures, and datasets while preserving the efficiency of plain mask transformers.

\subsection{Hyperparameter Sensitivity}

\begin{figure}[!t]
    \centering
    \includegraphics[width=0.49\columnwidth]{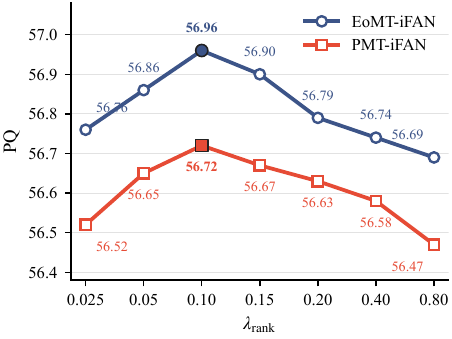}
    \hfill
    \includegraphics[width=0.49\columnwidth]{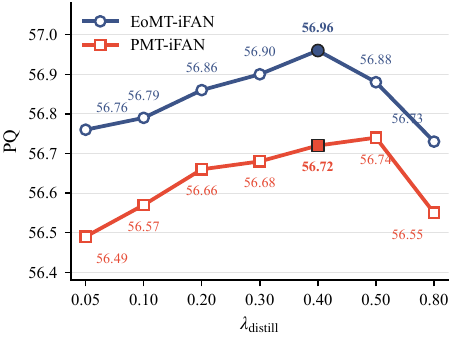}
    \caption{\textbf{Hyperparameter sensitivity.}
    Left: sensitivity to the APMR weight $\lambda_{\mathrm{rank}}$.
    Right: sensitivity to the CLSD weight $\lambda_{\mathrm{distill}}$.
    Each loss weight is varied independently with ViT-L and $640^2$ inputs, while all other settings are fixed.
    Filled markers indicate the selected values.}
    \label{fig:hyperparameter_sensitivity}
\end{figure}

Figure~\ref{fig:hyperparameter_sensitivity} evaluates the sensitivity of iFAN with EoMT-L and PMT-L to the loss weights of the APMR and CLSD objectives. Both models remain robust across a broad range of hyperparameter values.
We therefore adopt representative values within the stable regions, namely $\lambda_{\mathrm{rank}}=0.10$ and
$\lambda_{\mathrm{distill}}=0.40$, and keep them fixed across all datasets, backbones, input resolutions, and segmentation tasks.

\subsection{Ablation Studies}

\paragraph{Overall Ablation.}
Table~\ref{tab:overall_cld_component_ablation}(a) isolates the global mask-quality score, APMR, and CLSD. The EoMT baseline obtains 56.00 PQ. 
Adding mask-quality learning yields 56.43 PQ; further introducing ranking raises performance to 56.68 PQ, whereas combining quality learning with CLSD obtains 56.54 PQ. 
The complete iFAN model achieves 57.00 PQ. 
Thus, ranking suppresses competitive unmatched queries within the final prediction set, while CLSD transfers better adjusted probability-mask responses from intermediate layers to the final layer.

\begin{table}[t]
\centering
{%
\fontsize{9pt}{10.8pt}\selectfont
\rmfamily
\setlength{\tabcolsep}{0.9mm}
\renewcommand{\arraystretch}{1.08}

\begin{tabular*}{\columnwidth}{@{\extracolsep{\fill}}cc@{}}

\begin{tabular}[t]{@{}c@{}}
\textbf{(a) Overall module ablation} \\[2pt]
\begin{tabular}{cccc}
\toprule
Quality & APMR & CLSD & PQ \\
\midrule
$\times$     & $\times$     & $\times$     & 56.00 \\
$\checkmark$ & $\times$     & $\times$     & 56.43 \\
$\checkmark$ & $\checkmark$ & $\times$     & 56.68 \\
$\checkmark$ & $\times$     & $\checkmark$ & 56.54 \\
$\checkmark$ & $\checkmark$ & $\checkmark$ & \textbf{57.00} \\
\bottomrule
\end{tabular}
\end{tabular}

&

\begin{tabular}[t]{@{}c@{}}
\textbf{(b) CLSD component analysis} \\[2pt]
\begin{tabular}{cccc}
\toprule
$\mathcal{L}_{\mathrm{distill}}$ & BT & $\hat{S}$ Distill. & PQ \\
\midrule
$\times$     & $\times$     & $\times$     & 56.00 \\
$\checkmark$ & $\times$     & $\times$     & 55.96 \\
$\checkmark$ & $\checkmark$ & $\times$     & 56.39 \\
$\checkmark$ & $\times$     & $\checkmark$ & 56.02 \\
$\checkmark$ & $\checkmark$ & $\checkmark$ & \textbf{56.54} \\
\bottomrule
\end{tabular}
\end{tabular}

\end{tabular*}
}
\caption{\textbf{Overall iFAN and CLSD
ablations}
on COCO
\emph{val2017} using ViT-L at $640^2$. In (a), Quality denotes
the global mask-quality branch. In (b), a
checkmark under BT selects the intermediate teacher with the highest
soft-IoU (Eq.~\eqref{eq:teacher_selection}), whereas a cross fixes the
teacher to the layer immediately preceding the final layer. A checkmark
under $\hat{S}$ Distill. denotes distillation of the adjusted
probability-mask score map $\hat{S}$, whereas a cross distills
soft-IoU.}
\label{tab:overall_cld_component_ablation}
\end{table}

\paragraph{Ablation on Cross-Layer Self-Distillation.}
Table~\ref{tab:overall_cld_component_ablation}(b) studies best-teacher selection (BT) and $\hat{S}$-map distillation. 
Without self-distillation, the model obtains 56.00 PQ. 
Naively distilling soft-IoU from the preceding layer reduces the result to 55.96 PQ, indicating that the immediately preceding layer is not always a reliable teacher.
Selecting the intermediate teacher with the highest soft-IoU according to Eq.~\eqref{eq:teacher_selection} increases the result to 56.39 PQ, while using the adjusted probability-mask score map $\hat{S}$ with the preceding-layer teacher gives 56.02 PQ. 
Combining best-teacher selection with $\hat{S}$-map distillation reaches 56.54 PQ, 0.58 points above naive distillation and 0.15 points above best-teacher selection alone. 
Teacher selection therefore removes the main source of noisy cross-layer supervision, and $\hat{S}$-map distillation provides an additional benefit by transferring the same score map used for inference-time competition.

\begin{figure}[t]
    \centering
    \includegraphics[width=0.49\columnwidth]{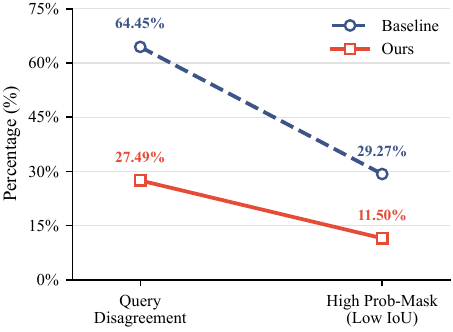}
    \hfill
    \includegraphics[width=0.49\columnwidth]{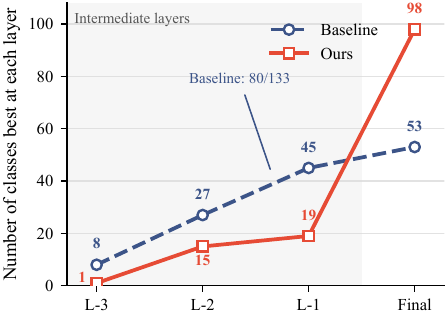}
    
    \caption{\textbf{Inference diagnostics on COCO \emph{val2017}.} 
    Left: disagreement between the \emph{max-prob-mask} and \emph{max-IoU} queries, together with the
    frequency of high probability-mask scores paired with low IoU.
    Right: the numbers of classes whose optimal prediction is produced by an intermediate or the final layer. Both panels compare EoMT with iFAN.}
    \label{fig:misalignment_diagnostics}
\end{figure}

Figure~\ref{fig:misalignment_diagnostics} (right) provides a depth-wise diagnostic. For EoMT, the final layer is optimal for 53 of the 133 classes, whereas an intermediate layer is optimal for the remaining 80. 
With CLSD, the number of final-layer-optimal classes increases to 98 and the intermediate-layer count decreases to 35. 
This shift indicates that CLSD more effectively preserves useful intermediate predictions in the final layer.

\begin{table}[t]
\centering
{%
\fontsize{9pt}{10.8pt}\selectfont
\rmfamily
\setlength{\tabcolsep}{1mm}
\renewcommand{\arraystretch}{1.0}

\begin{tabular}{@{}c@{\hspace{3mm}}c@{}}

\begin{tabular}[t]{@{}c@{}}
\textbf{(a) APMR component analysis} \\[2pt]
\begin{tabular}{cccc}
\toprule
$\mathcal{L}_{\mathrm{rank}}$ & $\hat{S}$ Ranking & SG & PQ \\
\midrule
$\times$     & $\times$     & $\times$     & 56.00 \\
$\checkmark$ & $\times$     & $\times$     & 56.39 \\
$\checkmark$ & $\checkmark$ & $\times$     & 56.45 \\
$\checkmark$ & $\checkmark$ & $\checkmark$ & \textbf{56.68} \\
\bottomrule
\end{tabular}
\end{tabular}

&

\begin{tabular}[t]{@{}c@{}}
\textbf{(b) Effect of model size} \\[2pt]
\begin{tabular}{lrrr}
\toprule
Size & Params & EoMT & iFAN \\
\midrule
ViT-S & 24M   & 44.7 & \textbf{46.3} \\
ViT-B & 93M   & 50.6 & \textbf{51.8} \\
ViT-L & 316M  & 56.0 & \textbf{57.0} \\
ViT-G & 1164M & 57.0 & \textbf{57.8} \\
\bottomrule
\end{tabular}
\end{tabular}

\end{tabular}
}

\caption{\textbf{APMR and model-size ablations} on COCO.
The variants in (a) cumulatively enable the APMR loss,
its computation on the adjusted probability-mask score map
$\hat{S}_i$, and stop-gradient (SG) on the class-probability
and mask branches. Panel (a) uses EoMT with ViT-L at $640^2$;
panel (b) uses $640^2$ inputs and DINOv2 pre-training and
reports PQ.}
\label{tab:cqr_model_size_ablation}
\end{table}

\paragraph{Ablation on Adjusted Probability-Mask Ranking.}
Table~\ref{tab:cqr_model_size_ablation}(a) incrementally evaluates the ranking design. 
Adding the ranking objective to the 56.00-PQ baseline reaches 56.39 PQ. 
Computing the ranking loss on the adjusted probability-mask score map $\hat{S}$ further improves PQ to 56.45. 
Finally, stopping the ranking gradients to the class-probability and mask branches, so that this objective updates the global quality score, yields 56.68 PQ. 
This final 0.23-point gain supports the design choice in the APMR section: ranking should calibrate the quality used in segmentation competition without disrupting the original classification and mask objectives.

The left panel of Figure~\ref{fig:misalignment_diagnostics} is consistent with the ablation. 
iFAN reduces the disagreement rate between the \emph{max-prob-mask} and \emph{max-IoU} queries from 64.45\% to 27.49\%, and reduces the proportion of high probability-mask scores with low IoU from 29.27\% to 11.50\%. 
APMR therefore makes the query selected for inference more consistent with the query that best overlaps the target.

\paragraph{Effect of model size.}
Table~\ref{tab:cqr_model_size_ablation}(b) evaluates EoMT-iFAN from ViT-S to ViT-G. iFAN improves all backbones by 1.6, 1.2, 1.0, and 0.8 PQ, respectively. The gain decreases as the backbone scales, suggesting that smaller models leave more probability-mask competition errors for iFAN to correct; the improvement remains consistent even for ViT-G.

\section{Conclusion}
We presented iFAN, an inference-aware learning framework for plain mask transformers that improves both query competition and cross-layer knowledge transfer. 
APMR augments the probability-mask score with global mask quality and explicitly ranks matched queries above hard unmatched competitors, making inference-time query selection more consistent with mask quality. 
CLSD identifies a better intermediate prediction for each target and distills its adjusted probability-mask score map to the final layer, without requiring multi-layer inference. Experiments on COCO, ADE20K, and Cityscapes show consistent improvements across panoptic, instance, and semantic segmentation, two plain mask-transformer baselines, multiple backbone sizes, and different input resolutions.
Diagnostic results further show that iFAN reduces the disagreement between the max-prob-mask and max-IoU queries from 64.45\% to 27.49\%, while increasing the number of classes for which the final layer achieves the best performance from 53 to 98. These substantial improvements provide direct evidence that iFAN effectively alleviates cross-query competition and preserves valuable intermediate information in the final predictions.

\bibliography{aaai2027}

\appendix
\clearpage

\twocolumn[
\begingroup
\let\Large\LARGE
\section{Supplementary Material}
\label{sec:supplementary}
\endgroup
]

\begin{table*}[!thb]
  \centering
  \caption{\textbf{Detailed panoptic segmentation results for PMT and
  EoMT on COCO \emph{val2017} and ADE20K \emph{val}.}
  $\mathrm{th}$ and $\mathrm{st}$ denote things and stuff,
  respectively. $\Delta\mathrm{PQ}$ compares each iFAN model with its
  matched baseline. Blue rows denote iFAN. Parameters, GFLOPs, and FPS
  in iFAN rows follow the matched baseline under shared-configuration
  accounting; ``--'' indicates that a metric is unavailable.}
  \label{tab:supp_detailed_panoptic}
  \resizebox{\textwidth}{!}{%
  \begin{tabular}{lllrrrrrrrrrrrrrr}
    \toprule
    Method & Dataset & Backbone & Input & Params & GFLOPs & FPS &
    PQ & RQ & SQ &
    PQ$_{\rm th}$ & RQ$_{\rm th}$ & SQ$_{\rm th}$ &
    PQ$_{\rm st}$ & RQ$_{\rm st}$ & SQ$_{\rm st}$ &
    $\Delta$PQ \\
    \midrule

    PMT & COCO & ViT-L & $640^2$ & 357M & 767 & 141 &
    56.1 & 67.0 & 82.9 & 61.5 & 73.0 & 83.6 &
    48.1 & 58.0 & 81.7 & -- \\
    \rowcolor[RGB]{230,245,255}
    PMT-iFAN & COCO & ViT-L & $640^2$ & 357M & 767 & 141 &
    \textbf{56.7} & 67.2 & 83.5 & 62.3 & 73.5 & 84.2 &
    48.2 & 57.6 & 82.4 & \textbf{+0.6} \\

    PMT & COCO & ViT-L & $1280^2$ & 357M & 4925 & 29 &
    58.1 & 68.7 & 83.7 & 64.6 & 75.7 & 84.8 &
    48.3 & 58.1 & 81.9 & -- \\
    \rowcolor[RGB]{230,245,255}
    PMT-iFAN & COCO & ViT-L & $1280^2$ & 357M & 4925 & 29 &
    \textbf{58.6} & 69.0 & 84.1 & 65.0 & 76.0 & 85.1 &
    48.9 & 58.4 & 82.7 & \textbf{+0.5} \\

    EoMT & COCO & ViT-L & $640^2$ & 316M & 669 & 128 &
    56.0 & 67.2 & 82.5 &
    61.2 & 73.1 & 83.1 &
    48.2 & 58.2 & 81.5 & -- \\
    \rowcolor[RGB]{230,245,255}
    EoMT-iFAN & COCO & ViT-L & $640^2$ & 316M & 669 & 128 &
    \textbf{57.0} & 68.0 & 83.0 &
    62.1 & 73.7 & 83.6 &
    49.3 & 59.3 & 82.1 & \textbf{+1.0} \\

    EoMT & COCO & ViT-L & $1280^2$ & 322M & 4146 & 30 &
    58.3 & 69.0 & 83.5 &
    64.7 & 75.9 & 84.7 &
    48.6 & 58.6 & 81.8 & -- \\
    \rowcolor[RGB]{230,245,255}
    EoMT-iFAN & COCO & ViT-L & $1280^2$ & 322M & 4146 & 30 &
    \textbf{58.8} & 69.5 & 83.8 &
    65.2 & 76.5 & 84.7 &
    49.2 & 58.9 & 82.4 & \textbf{+0.5} \\

    EoMT & COCO & ViT-G & $640^2$ & 1164M & 2261 & 55 &
    57.0 & 68.2 & 82.7 &
    62.6 & 74.5 & 83.5 &
    48.6 & 58.8 & 81.4 & -- \\
    \rowcolor[RGB]{230,245,255}
    EoMT-iFAN & COCO & ViT-G & $640^2$ & 1164M & 2261 & 55 &
    \textbf{57.8} & 68.8 & 83.2 &
    63.5 & 75.4 & 83.8 &
    49.1 & 58.8 & 82.4 & \textbf{+0.8} \\

    EoMT & COCO & ViT-G & $1280^2$ & 1171M & 12712 & 12 &
    59.2 & 69.8 & 83.8 &
    66.0 & 77.2 & 85.1 &
    48.8 & 58.8 & 81.8 & -- \\
    \rowcolor[RGB]{230,245,255}
    EoMT-iFAN & COCO & ViT-G & $1280^2$ & 1171M & 12712 & 12 &
    \textbf{59.6} & 70.4 & 83.9 &
    66.3 & 77.6 & 85.0 &
    49.6 & 59.5 & 82.3 & \textbf{+0.4} \\

    \midrule

    PMT & ADE20K & ViT-L & $640^2$ & 357M & 767 & 141 &
    49.4 & 58.9 & 82.7 & 48.3 & 57.9 & 82.3 &
    52.0 & 60.8 & 83.5 & -- \\
    \rowcolor[RGB]{230,245,255}
    PMT-iFAN & ADE20K & ViT-L & $640^2$ & 357M & 767 & 141 &
    \textbf{50.7} & -- & -- & 50.4 & -- & -- &
    51.2 & -- & -- & \textbf{+1.3} \\

    PMT & ADE20K & ViT-L & $1280^2$ & 357M & 4925 & 29 &
    50.5 & 59.6 & 83.2 & 50.5 & 59.6 & 83.7 &
    51.4 & 59.6 & 82.3 & -- \\
    \rowcolor[RGB]{230,245,255}
    PMT-iFAN & ADE20K & ViT-L & $1280^2$ & 357M & 4925 & 29 &
    \textbf{53.0} & -- & -- & 53.7 & -- & -- &
    51.6 & -- & -- & \textbf{+2.5} \\

    EoMT & ADE20K & ViT-L & $640^2$ & 316M & 669 & 128 &
    50.6 & 59.9 & 82.4 &
    49.7 & 59.2 & 82.2 &
    52.3 & 61.5 & 82.8 & -- \\
    \rowcolor[RGB]{230,245,255}
    EoMT-iFAN & ADE20K & ViT-L & $640^2$ & 316M & 669 & 128 &
    \textbf{52.2} & 62.0 & 81.3 &
    51.8 & 61.9 & 81.9 &
    53.1 & 62.1 & 80.1 & \textbf{+1.6} \\

    EoMT & ADE20K & ViT-L & $1280^2$ & 322M & 4146 & 30 &
    51.7 & 60.4 & 82.5 &
    52.0 & 61.2 & 82.5 &
    51.0 & 58.9 & 82.6 & -- \\
    \rowcolor[RGB]{230,245,255}
    EoMT-iFAN & ADE20K & ViT-L & $1280^2$ & 322M & 4146 & 30 &
    \textbf{53.9} & 63.2 & 84.6 &
    53.7 & 63.3 & 84.1 &
    54.1 & 62.9 & 85.5 & \textbf{+2.2} \\

    EoMT & ADE20K & ViT-G & $640^2$ & 1164M & 2261 & 55 &
    51.3 & 60.7 & 83.3 &
    50.5 & 59.9 & 82.6 &
    53.0 & 62.3 & 84.6 & -- \\
    \rowcolor[RGB]{230,245,255}
    EoMT-iFAN & ADE20K & ViT-G & $640^2$ & 1164M & 2261 & 55 &
    \textbf{52.7} & 62.3 & 83.2 &
    52.3 & 62.2 & 83.1 &
    53.6 & 62.7 & 83.4 & \textbf{+1.4} \\

    EoMT & ADE20K & ViT-G & $1280^2$ & 1171M & 12712 & 12 &
    52.8 & 61.3 & 84.0 &
    52.7 & 61.6 & 83.9 &
    52.6 & 60.8 & 84.1 & -- \\
    \rowcolor[RGB]{230,245,255}
    EoMT-iFAN & ADE20K & ViT-G & $1280^2$ & 1171M & 12712 & 12 &
    \textbf{53.9} & 63.0 & 84.3 &
    54.0 & 63.2 & 83.8 &
    53.7 & 62.6 & 85.1 & \textbf{+1.1} \\

    \bottomrule
  \end{tabular}%
  }
\end{table*}

\begin{table*}[!t]
  \centering
  \caption{\textbf{Detailed semantic segmentation results for PMT and
  EoMT on Cityscapes and ADE20K \emph{val}.}
 Block $-k$ gives the
  validation mIoU predicted from the $k$-th block counted backward from
  the final block.
  $\Delta$mIoU compares each iFAN model with its matched baseline. Blue
  rows denote iFAN, and ``--'' indicates that a metric was not logged
  for that configuration.}
  \label{tab:supp_detailed_semantic}
  \resizebox{\textwidth}{!}{%
  \begin{tabular}{lllrrrrrrrrrr}
    \toprule
    Method & Dataset & Backbone & Input & Params & GFLOPs & FPS &
    mIoU  & Block $-1$ & Block $-2$ & Block $-3$ & Block $-4$ &
    $\Delta$mIoU \\
    \midrule

    EoMT & Cityscapes & ViT-L & $1024^2$ & 319M & 4350 & 25 &
    84.2  & -- & -- & -- & -- & -- \\
    \rowcolor[RGB]{230,245,255}
    EoMT-iFAN & Cityscapes & ViT-L & $1024^2$ & 319M & 4350 & 25 &
    \textbf{84.5} &
    84.5   & 84.4 &82.6 & 74.3 & \textbf{+0.3} \\

    \midrule

    PMT & ADE20K & ViT-L & $512^2$ & 357M & 823 & 128 &
    58.5  & -- & -- & -- & -- & -- \\
    \rowcolor[RGB]{230,245,255}
    PMT-iFAN & ADE20K & ViT-L & $512^2$ & 357M & 823 & 128 &
    \textbf{59.4}  & -- & -- & -- & -- &
    \textbf{+0.9} \\

    EoMT & ADE20K & ViT-L & $512^2$ & 319M & 721 & 92 &
    58.4  & -- & -- & -- & -- & -- \\
    \rowcolor[RGB]{230,245,255}
    EoMT-iFAN & ADE20K & ViT-L & $512^2$ & 319M & 721 & 92 &
    \textbf{59.0}  &
    58.5 & 58.2 & 57.5 & 14.1 & \textbf{+0.6} \\

    \bottomrule
  \end{tabular}%
  }
\end{table*}

\begin{table*}[!t]
  \centering
  \caption{\textbf{Detailed local instance segmentation results for
  PMT and EoMT on COCO \emph{val2017}.}
   $\Delta$AP compares each iFAN model with its matched
  baseline; EoMT deltas are computed from the unrounded records before
  rounding. Blue rows denote iFAN. Parameters and FPS in iFAN rows follow the
  matched baseline under shared-configuration accounting.}
  \label{tab:supp_detailed_instance}
  \resizebox{\textwidth}{!}{%
  \begin{tabular}{llrrrrrrrrrrrr}
    \toprule
    Method & Backbone & Input & Params & FPS &
    AP & AP$_{50}$ & AP$_{75}$ & AP$_S$ & AP$_M$ & AP$_L$ &
    Epoch & $\Delta$AP \\
    \midrule

    PMT & ViT-L & $640^2$ & 357M & 141 &
    45.4 & 69.5 & 48.2 & 21.0 & 50.9 & 72.3 & 11  & -- \\
    \rowcolor[RGB]{230,245,255}
    PMT-iFAN & ViT-L & $640^2$ & 357M & 141 &
    \textbf{46.6} & 69.8 & 50.2 & 21.7 & 51.2 & 72.5 &
    10  & \textbf{+1.2} \\

    EoMT & ViT-L & $640^2$ & 316M & 128 &
    44.8 & 69.3 & 47.4 & 19.7 & 50.6 & 72.2 &
    -- & -- \\
    \rowcolor[RGB]{230,245,255}
    EoMT-iFAN & ViT-L & $640^2$ & 316M & 128 &
    \textbf{46.0} & 69.7 & 48.9 &
    20.4 & 50.6 & 71.9 & --  & \textbf{+1.2} \\

    \midrule

    PMT & ViT-L & $1280^2$ & 357M & 29 &
    49.3 & 73.5 & 53.5 & 28.7 & 53.9 & 72.9 & 11  & -- \\
    \rowcolor[RGB]{230,245,255}
    PMT-iFAN & ViT-L & $1280^2$ & 357M & 29 &
    \textbf{50.6} & 74.0 & 55.1 & 28.8 & 54.6 & 72.8 &
    10 &  \textbf{+1.3} \\

    EoMT & ViT-L & $1280^2$ & 322M & 30 &
    48.8 & 73.5 & 52.5 & 27.4 & 54.1 & 72.1 &
    -- &  -- \\
    \rowcolor[RGB]{230,245,255}
    EoMT-iFAN & ViT-L & $1280^2$ & 322M & 30 &
    \textbf{50.6} & 74.2 & 54.8 &
    29.6 & 54.7 & 72.9 & -- & \textbf{+1.8} \\

    \bottomrule
  \end{tabular}%
  }
\end{table*}

\subsection{More ablation}
\label{sec:more_ablation}

\begin{figure}[!h]
    \centering
    \begin{minipage}[t]{0.49\columnwidth}
        \centering
        \includegraphics[width=\linewidth]
        {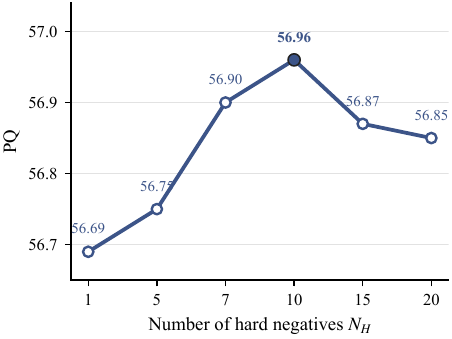}
    \end{minipage}\hfill
    \begin{minipage}[t]{0.49\columnwidth}
        \centering
        \includegraphics[width=\linewidth]
        {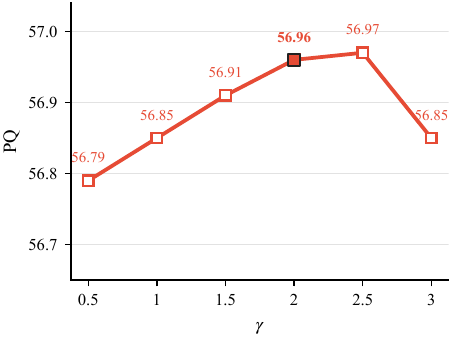}
    \end{minipage}
    \caption{\textbf{Hyperparameter sensitivity on COCO panoptic segmentation.}
    Left: performance remains stable across a broad range of $N_H$.
    Right: $\gamma$ balances the suppression of low-valued responses against
    the preservation of useful responses.
    Filled markers denote the selected settings,
    $N_H=10$ and $\gamma=2$.}
    \label{fig:hard_negative_gamma_sensitivity}
\end{figure}

Figure~\ref{fig:hard_negative_gamma_sensitivity} evaluates the sensitivity to
the number of hard negatives $N_H$ and the weighting exponent $\gamma$.
Performance remains consistently strong for $N_H\in[7,15]$.
Similarly, varying $\gamma$ from 0.5 to 3 results in only a 0.18 PQ difference.
The setting $\gamma=2$ effectively suppresses low-valued responses while
avoiding the excessive attenuation caused by a larger exponent.
We therefore use $N_H=10$ and $\gamma=2$ as the default settings.

\subsection{Detailed PMT and EoMT Results}
\label{sec:supp_pmt_results}

Tables~\ref{tab:supp_detailed_panoptic}--\ref{tab:supp_detailed_instance}
We provide more detailed results of EoMT, PMT, and iFAN for panoptic, semantic, and instance segmentation.
For the PMT ViT-L 1280 results reported in the main paper, there was a transcription error. The correct results are the ones provided in the appendix.

\subsection{Score--Quality Alignment in Query Competition}
\label{sec:appendix_query_competition}

This section details the query-wise diagnostic summarized in Fig.1 (top) of the Introduction. 
It tests whether the query-level score that determines probability-mask competition is consistent with the actual mask quality of that query, thereby isolating the inference mismatch targeted by Adjusted Probability-Mask Ranking (APMR).

\paragraph{Protocol.}
We evaluate EoMT and EoMT-iFAN on all 5,000 images of COCO
\emph{val2017}. For a given image, let
$\{(c_j,M_j)\}_{j=1}^{N_{\mathrm{gt}}}$ denote its ground-truth
segments, where $c_j$ and $M_j$ are the class label and binary mask of
segment $j$. Query $i$ outputs a class-probability distribution
$\hat{\mathbf{p}}_i$ over the foreground classes and the no-object
class $\varnothing$, together with a soft mask
$\hat{M}_i\in[0,1]^{H\times W}$. Let
$\hat{c}_i=\arg\max_c\hat{\mathbf{p}}_i(c)$ denote its predicted class.
We first compute its soft-IoU with every ground-truth segment:
\begin{equation}
 s_{ij}=\operatorname{\textit{soft\_IoU}}(\hat{M}_i,M_j)\\
\label{eq:supp_soft_iou}
\end{equation}
The sums are taken over all image pixels; unlike thresholded IoU,
Eq.~\eqref{eq:supp_soft_iou} retains the predicted mask probabilities.

To exclude queries with essentially no spatial support on any annotated
segment, we define the evaluated query set as
\begin{equation}
 \mathcal{Q}_{\mathrm{eval}}
 =
 \left\{
 i\,\middle|\,
 \hat{c}_i\neq\varnothing,\ 
 \max_{1\leq j\leq N_{\mathrm{gt}}}s_{ij}>0.01
 \right\}.
\label{eq:supp_query_filter}
\end{equation}
This initial overlap test is class-agnostic and serves only as a
candidate filter. It does not select the ground-truth segment used to
define mask quality.

For each $i\in\mathcal{Q}_{\mathrm{eval}}$, let
$\mathcal{J}_i=\{j\mid c_j=\hat{c}_i\}$ be the set of ground-truth
segments whose class matches the predicted class of query $i$. We
define the query's reference \textbf{mask quality} as
\begin{equation}
 u_i =
 \begin{cases}
 \displaystyle\max_{j\in\mathcal{J}_i}s_{ij},
 & \mathcal{J}_i\neq\varnothing,\\[3pt]
 0, & \mathcal{J}_i=\varnothing.
 \end{cases}
\label{eq:supp_reference_quality}
\end{equation}
Hence, $u_i$ is the best soft-IoU that query $i$ attains with any
ground-truth segment of its predicted class. Overlap with a
ground-truth segment of a different class cannot increase $u_i$.
Unlike one-to-one Hungarian assignment, this diagnostic comparison is
many-to-one: multiple queries may use the same ground-truth segment as
their best class-consistent reference. It therefore measures the
class-consistent mask quality of every retained competing query without
tying the analysis to its training-time assignment.

For EoMT, the query-level \textbf{competition score} is the predicted-class confidence
\begin{equation}
r_i^{\mathrm{EoMT}} =
\max_{c\neq\varnothing}\hat{\mathbf{p}}_i(c).
\end{equation}
For iFAN, the corresponding score incorporates the predicted global
mask quality,
\begin{equation}
r_i^{\mathrm{iFAN}} =
\hat{q}_i\,
\max_{c\neq\varnothing}\hat{\mathbf{p}}_i(c).
\end{equation}
The adjusted probability-mask score map used by iFAN is therefore $\hat{S}_i=(r_i^{\mathrm{iFAN}})^2\hat{M}_i$. Because squaring is monotonic for these non-negative scores, it does not change the pairwise ordering or the identity of the highest-scoring query.

\paragraph{Metrics.}
We report five complementary measurements:

\smallskip
\noindent\textbf{Pairwise ordering accuracy (higher is better):}
For every pair of retained queries from the same image with
$|u_i-u_k|>0.05$, this metric checks whether their competition scores
have the same ordering as their reference mask qualities. It reports
the percentage of such pairs for which the higher-quality query also
has the higher score.

\smallskip
\noindent\textbf{Top-query accuracy (higher is better):}
For each eligible image, we identify the retained query with the
largest competition score and the retained query with the largest
reference mask quality. This metric reports the percentage of images
for which these two selections identify the same query.

\smallskip
\noindent\textbf{Top-query conflict (lower is better):}
Let $i_r=\arg\max_{i\in\mathcal{Q}_{\mathrm{eval}}}r_i$ be the
highest-scoring retained query. A conflict is counted when its quality
is more than 0.05 below the best available quality, i.e.,
$\max_{i\in\mathcal{Q}_{\mathrm{eval}}}u_i-u_{i_r}>0.05$. The metric
is the percentage of eligible images containing such a conflict.

\smallskip
\noindent\textbf{High-score low-quality rate (lower is better):}
Among retained queries with $r_i\geq0.8$, this metric reports the
percentage whose reference mask quality satisfies $u_i<0.5$. It
measures how often a highly confident competitor nevertheless has poor
class-consistent mask quality.

\smallskip
\noindent\textbf{Mask-quality gap (lower is better):}
For each eligible image, we compute
$\max_{i\in\mathcal{Q}_{\mathrm{eval}}}u_i-u_{i_r}$, namely the
quality lost by selecting the highest-scoring query instead of the
best-quality retained query. We report its average over eligible
images; a value of zero indicates no quality loss.

\begin{table}[t]
\centering
\small
\caption{\textbf{Score--quality alignment on COCO \emph{val2017}.}
All images are evaluated. ``Change'' denotes iFAN minus EoMT; percentage
changes are in percentage points.}
\label{tab:query_competition_diagnostic}
\begin{tabularx}{\columnwidth}{@{}Xrrr@{}}
\toprule
Metric & EoMT & iFAN & Change \\
\midrule
Pairwise ordering accuracy (\%) & 76.07 & \textbf{79.42} & +3.35 \\
Top-query accuracy (\%)         & 35.55 & \textbf{72.51} & +36.96 \\
Top-query conflict (\%)         & 64.45 & \textbf{27.49} & -36.96 \\
High-score low-quality rate (\%)& 29.27 & \textbf{11.5}  & -17.8 \\
Mask-quality gap                & 0.1070 & \textbf{0.0591} & -0.0479 \\
\bottomrule
\end{tabularx}
\end{table}

\paragraph{Results and interpretation.}

Pairwise ordering accuracy increases from 76.07\% to 79.42\%. 
The effect is larger at the top of the ranking: top-query accuracy increases from 35.55\% to 72.51\%, top-query conflict decreases from 64.45\% to 27.49\%, and the average mask-quality gap decreases from 0.1070 to 0.0591. 
The high-score low-quality rate also decreases from 29.27\% to approximately 11.5\%. APMR therefore improves the high-score region that determines which query dominates probability-mask competition. 
The statistics contain 72,619 valid queries and 1,474 eligible images for EoMT, and 72,209 queries and 1,473 images for iFAN; the small count difference results from applying the candidate criterion independently to each model.

\subsection{Qualitative Segmentation Results}
\label{sec:supp_qualitative}

\begin{figure*}[p]
  \centering
  \includegraphics[
    width=\textwidth,
    height=0.86\textheight,
    keepaspectratio
  ]{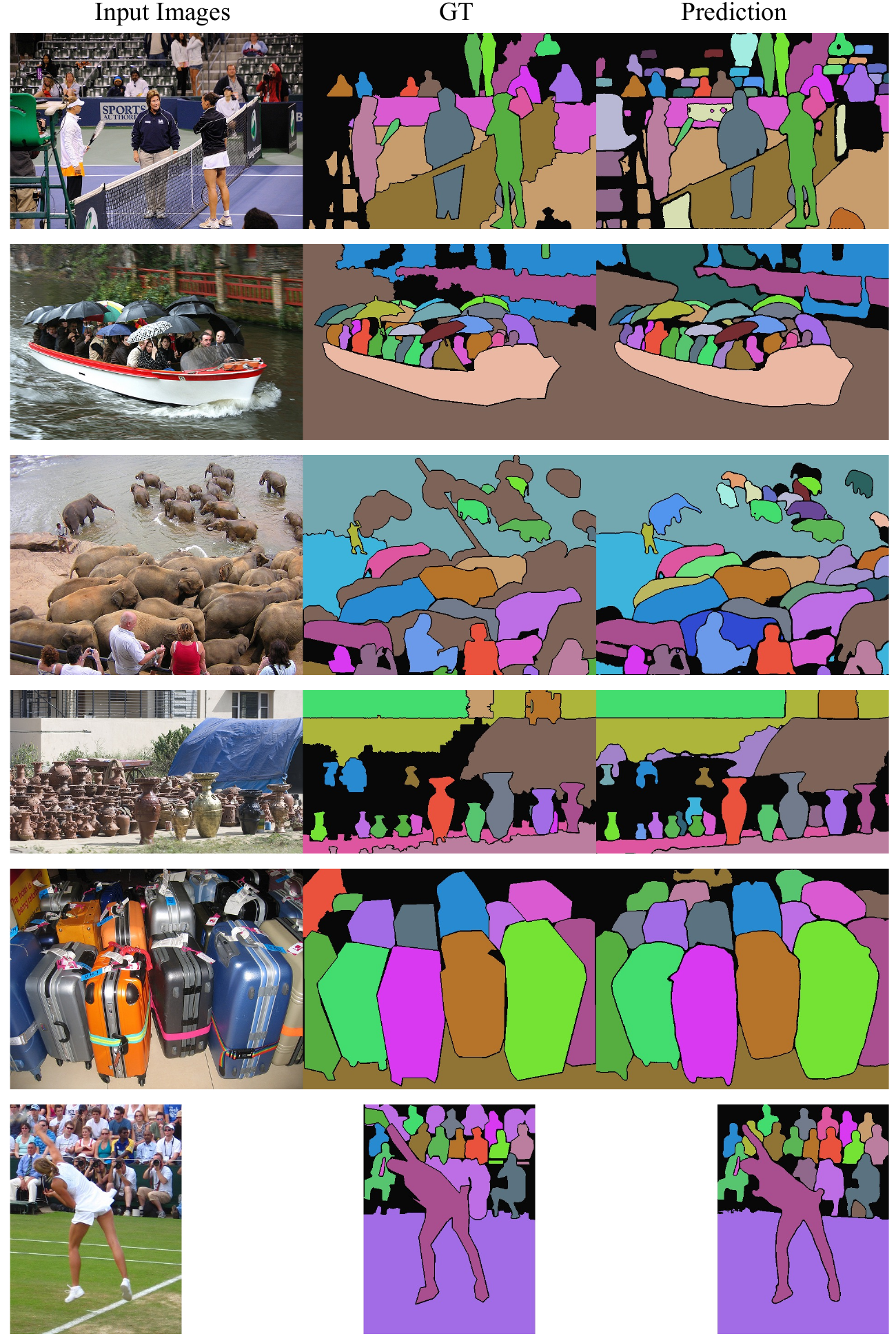}
  \caption{\textbf{Qualitative results for panoptic segmentation.}
  Each row shows an input image, ground truth, and the standalone prediction. The examples contain complex
  street layouts with large stuff regions and multiple thing
  instances, including vehicles, pedestrians, traffic signs, and
  roadside objects.}
  \label{fig:supp_panoptic}
\end{figure*}

\begin{figure*}[p]
  \centering
  \includegraphics[
    width=\textwidth,
    height=0.84\textheight,
    keepaspectratio
  ]{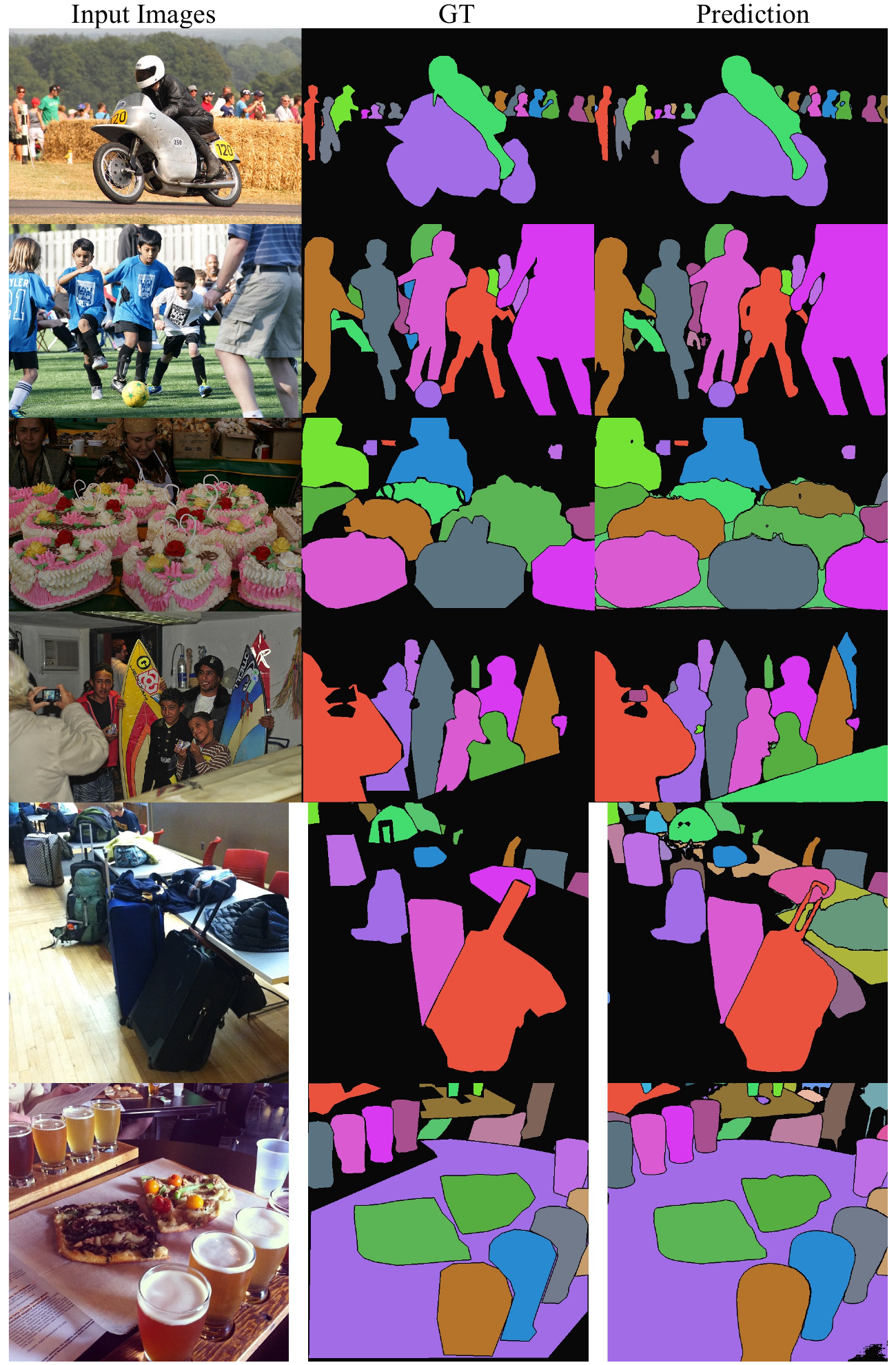}
  \caption{\textbf{Qualitative results for instance segmentation.}
  The three columns show the input image, the ground truth, and
  the standalone instance masks. The selected scenes emphasize dense
  arrangements, substantial occlusion, scale variation, and adjacent
  instances with similar appearance.}
  \label{fig:supp_instance}
\end{figure*}

\begin{figure*}[p]
  \centering
  \includegraphics[
    width=\textwidth,
    height=0.84\textheight,
    keepaspectratio
  ]{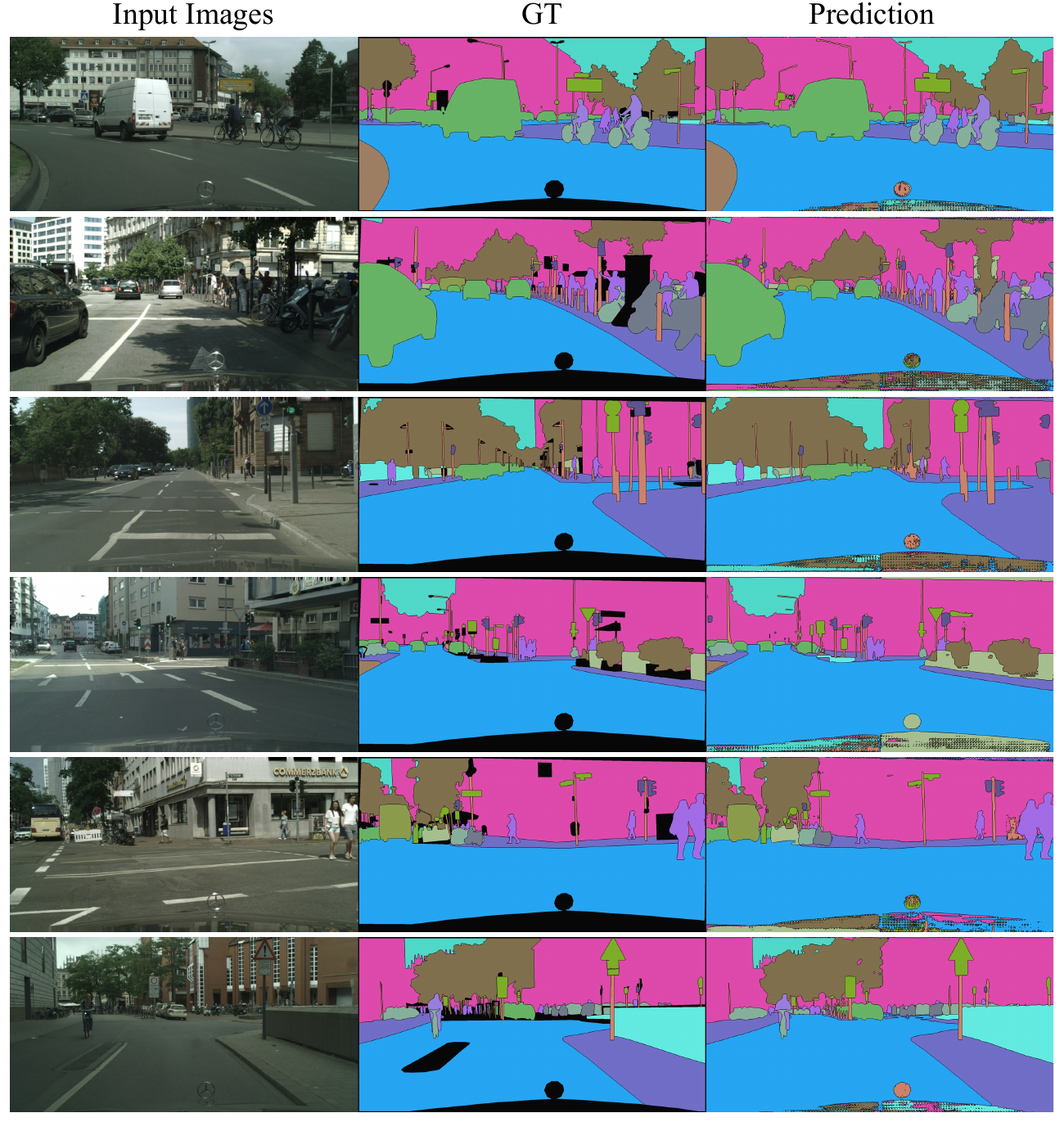}
  \caption{\textbf{Qualitative results for semantic segmentation.}
  Each row presents the input image, the ground truth, and the standalone prediction. The examples cover diverse
  indoor and outdoor scenes and illustrate coherent region assignment
  around objects with different shapes, scales, and surrounding
  context.}
  \label{fig:supp_semantic}
\end{figure*}

Figures~\ref{fig:supp_panoptic}, \ref{fig:supp_instance},
and~\ref{fig:supp_semantic} show representative iFAN predictions for
panoptic, instance, and semantic segmentation. Each row presents the
input image, ground truth, and standalone prediction. These
examples complement, rather than replace, the benchmark comparisons.

\end{document}